\documentclass[conference]{IEEEtran}

\usepackage{color}
\usepackage{lettrine}
\usepackage{multirow}
\usepackage{algorithm,algorithmic}
\newcommand{\rpm}{\sbox0{$1$}\sbox2{$\scriptstyle\pm$}
\raise\dimexpr(\ht0-\ht2)/2\relax\box2 }
\usepackage{graphicx}
\usepackage{booktabs}
\usepackage{adjustbox}
\usepackage{tabularx}
\usepackage{cite}

\usepackage{eso-pic}

\begin{document}
\AddToShipoutPicture*{\small \sffamily\raisebox{1.2cm}{\hspace{1.8cm}978-1-7281-0397-6/19/\$31.00 ©2019 IEEE}}

\title{Accelerating Deterministic and Stochastic Binarized Neural Networks on FPGAs Using OpenCL}
\author{\IEEEauthorblockN{Corey Lammie, Wei Xiang, and Mostafa Rahimi Azghadi}
	\IEEEauthorblockA{College of Science and Engineering, James Cook University, Queensland 4814, Australia\\ 
		Email:\{corey.lammie, wei.xiang, mostafa.rahimiazghadi\}@jcu.edu.au}}
\maketitle

\begin{abstract}
Recent technological advances have proliferated the available computing power, memory, and speed of modern Central Processing Units (CPUs), Graphics Processing Units (GPUs), and Field Programmable Gate Arrays (FPGAs). Consequently, the performance and complexity of Artificial Neural Networks (ANNs) is burgeoning. While GPU-accelerated Deep Neural Networks (DNNs) currently offer state-of-the-art performance, they consume large amounts of power. Training such networks on CPUs is inefficient, as data throughput and parallel computation is limited. FPGAs are considered a suitable candidate for performance critical, low power systems, e.g. the Internet of Things (IOT) edge devices. Using the Xilinx SDAccel or Intel FPGA SDK for OpenCL development environment, networks described using the high-level OpenCL framework can be accelerated on heterogeneous platforms. Moreover, the resource utilization and power consumption of DNNs can be further enhanced by utilizing regularization techniques that binarize network weights. In this paper, we introduce, to the best of our knowledge, the first FPGA-accelerated stochastically binarized DNN implementations, and compare them to implementations accelerated on both GPUs and FPGAs. 
All our developed networks are trained and benchmarked using the popular MNIST and CIFAR-10 datasets. For our binarized and conventional FPGA-based networks, we achieve a $>$16-fold improvement in power consumption, compared to their GPU-accelerated counterparts. Also, our binarized FPGA-based networks require $>$25\% shorter inference times, compared to their GPU-based counterparts. 
\end{abstract}

\IEEEpeerreviewmaketitle

\color{black}
\section{Introduction}
\lettrine{D}{eep} Neural Network (DNN) architectures have become integral to a variety of applications in Artificial Intelligence (AI) and Machine Learning (ML). While these learning networks and their underpinned elements have been actively researched since 1974~\cite{deep_learning_origin}, the inception of recent modern GPUs and faster CPU architectures have greatly facilitated Neural Network (NN) research and enabled the development of highly accurate and complex DNNs.  

However, high-performance CPU- and GPU-accelerated DNNs are putative to consume large amounts of power. As a result, accelerating DNNs on low-power and resource-constrained devices, such as portable smart electronics and IOT edge devices, becomes formidable. Considerable efforts are currently being made to utilize customized hardware solutions using FPGAs, presenting significant reductions in power consumption using both Fully Connected (FC) and Convolutional Neural Networks (CNNs) \cite{pipe_cnn, gpu_outperform_fpga,8702248}.

Despite the many improvements that recent FPGA studies offer in boosting parallelism and power efficiency, due to the large number of high-resolution multiplications required during learning and inference, such accelerated implementations are inhibited by the amount of dedicated multipliers and Digital Signal Processing (DSP) blocks available on FPGAs. Therefore, new techniques have recently been developed to account for the limited hardware resources available. A very popular technique, which quantizes network weights to binary states, has been proposed in order to greatly reduce resource utilization, and as a result, power consumption, while exhibiting minimal performance degradation~\cite{2015arXiv151100363C}. Within these networks, denoted \textit{Binarized Neural Networks (BNNs)}, many resource-hungry multiply-accumulate operations, required during learning and inference, are replaced with simple accumulations. 

Deterministic~\cite{PPM}, stochastic~\cite{2015arXiv151100363C}, and recursive~\cite{selfbinary} BNNs binarize weights during forward and backwards learning propagation cycles, while retaining precision of the stored weights to which gradients are accumulated. Self-binarizing networks~\cite{selfbinary} train using a unique representation of network weights, involving a smooth activation function, which is iteratively sharpened during training until it becomes a binary representation equivalent to the sign activation function.

While hardware implementations of deterministic BNNs are plentiful~\cite{gpu_outperform_fpga, Yang2018}, to the best of our knowledge, there are no current FPGA implementations of stochastic BNNs. Therefore, here we propose the first FPGA implementations of stochastic BNNs, as it has been demonstrated that stochastic BNNs further improve the learning performance of BNNs, compared to their deterministic counterparts~\cite{2015arXiv151100363C}.

In addition, we provide comprehensive results through investigating the acceleration of deterministic and stochastic BNNs on both GPUs and FPGAs using High Level Synthesis (HLS) techniques utilizing the OpenCL framework, to encourage deployment using heterogeneous platforms.
Resource usage and performance of the implemented networks are also compared for permutation-invariant DNNs, and CNNs, trained and tested for MNIST~\cite{mnist} and CIFAR-10~\cite{cifar}. For all our hardware implementations, we draw comparisons among designs utilizing deterministic, stochastic, or no regularization techniques. Our specific contributions are as follows:

\begin{itemize}
  \item We implement and present the first FPGA-accelerated stochastically binarized DNNs and CNNs.
  \item We employ complete FPGA-accelerated DNNs and CNNs on a standalone System On a Chip (SoC), requiring no host computer or extra device for partial computation.
  \item We demonstrate that our new binarized FPGA-accelerated DNNs and CNNs offer significantly reduced power usage and shorter inference times, compared to their equivalent full resolution counterparts, on MNIST and CIFAR-10, implemented on both GPU and FPGA.
  \item We report and investigate the learning times required for all of our implemented networks.  
\end{itemize}

\section{Preliminaries}
This section briefly reviews and presents the algorithms and methods used in our developed networks for the MNIST and CIFAR-10 classification benchmarks.

\begin{algorithm}[!b]
	\caption{Training Algorithm of the Accelerated Binarized Neural Networks}
	\begin{algorithmic}[1]
		\renewcommand{\algorithmicrequire}{\textbf{Input:}}
		\renewcommand{\algorithmicensure}{\textbf{Output:}}
		\REQUIRE a mini-batch of (inputs, targets), previous parameters $w_{t-1}$ and $b_{t-1}$, and a learning rate $\eta$.
		\ENSURE  updated parameters $w_t$ and $b_t$.
		\\\STATE \textbf{Forward Propogation}
		\\$w_b \leftarrow$ binarize $(w_{t-1})$.
		\\For $k = 1$ to $L$, compute $a_k$ knowing $a_{k-1}, w_b, b_{t-1}$.
		\\ \STATE \textbf{Backward Propogation}
		\\ Initialize output layer's activation gradient $\frac{\partial C}{\partial a_L}$
		\\ For $k = L$ to $2$, compute $\frac{\partial C}{\partial a_{k-1}}$ using $\frac{\partial C}{\partial a_{k}}$ and $w_b$.
		\\ \STATE \textbf{Parameter Update}
		\\ Compute $\frac{\partial C}{\partial w_b}$ and $\frac{\partial C}{\partial db_{t-1}}$, using $\frac{\partial C}{\partial a_k}$ and $a_{k-1}$.
		\\ $w_t \leftarrow $ clip$(w_{t-1} - \eta \frac{\partial C}{\partial w_b})$
		\\ $b_t \leftarrow b_{t-1} - \eta \frac{\partial C}{\partial b_{t-1}}$.		
		\\ \STATE \textbf{Weight Normalization}
		\\ $w \leftarrow \textnormal{clip}(w)$
		
	\end{algorithmic} \label{bnn_train_alg}
\end{algorithm}

\subsection{Binary Weight Regularization}
Binary weight regularization~\cite{2015arXiv151100363C}, constrains network weights to binary states of \{+1, -1\}, during forward and backward propagations. The binarization operation transforms the full-precision weights into binary values, using either a deterministic or a stochastic approach.

\subsubsection{Deterministic Binarization}
Deterministic binarization is defined in Equation (\ref{det_binarization}).

\begin{equation}\label{det_binarization}
w_b = \left\{\begin{array}{lr}
-1 & \textnormal{if } w \leq 0\\
+1 & \textnormal{otherwise},
\end{array}\right.
\end{equation}
where $w_b$ is the binarized weight and $w$ is the real-valued full-precision weight.

\color{black}
\subsubsection{Stochastic Binarization}
Stochastic binarization is an alternative binarization technique, which stochastically binarizes weights. The stochastic binary projection is presented in Equation (\ref{stochastic_bin}).

\begin{equation}\label{stochastic_bin}
w_b = \left\{\begin{array}{ll}
+1 & \textnormal{with probability } \rho = \sigma(w),\\
-1 & \textnormal{with probability } 1 - \rho,
\end{array}\right.
\end{equation}
where $\sigma$ is the hard sigmoid function described in Equation (\ref{hard_sigmoid}).

\begin{equation}\label{hard_sigmoid}
\sigma(x) = \textnormal{clip} (\frac{x+1}{2},0,1) = \textnormal{max} (0, \textnormal{min} (1,\frac{x+1}{2})).
\end{equation}

\subsection{Training Algorithm}
Algorithm (\ref{bnn_train_alg}) provides a high-level overview of the training algorithm used for deterministic and stochastic BNNs.
Here, $w$, $b$, and $\eta$ represent the weights, biases, and learning rate, while $C$ denotes the cost function for each mini-batch. Furthermore, $w_b$ represents binary weights and $a_k$ represents the $k$th layer activation function, while binarize() implements Equation (\ref{det_binarization}) or (\ref{stochastic_bin}) depending the utilized regularization, and clip() clips values between $-1$ and $+1$. By adopting this training algorithm, during learning and inference, network outputs can be determined using simple Multiply and Accumulate (MAC) operations, in-place of dedicated multiplier blocks~\cite{PPM}.

\color{black}
\section{Network Architecture}
The complete architecture of the implemented networks consists of two main components: \textit{A. The Software Architecture}, and \textit{B. The Hardware Architecture}. The software architecture defines the targeted neural network structure, which will be described in C++ and OpenCL kernels. The hardware architecture describes the integration between the hardware used to run the OpenCL kernels and a host controller, which is the program executed on a host processor. This processor is used to launch OpenCL kernels and to manage device memory.

\subsection{Software Architecture}
We implement two distinct neural network architectures: a permutation-invariant FC network for MNIST, and the VGG-16~\cite{vgg} CNN for CIFAR-10. Details pertaining to each network are provided in a publicly available GitHub repository\footnote{https://github.com/coreylammie/Accelerating-Stochastically-Binarized-Neural-Networks-on-FPGAs-using-OpenCL}.

To decrease the quantization error, which binarization introduces, the output of each layer is normalized using batch normalization. The output of the final layer is fed through a Softmax activation layer, and the network's loss is minimized using cross-entropy. SGD with momentum is used to optimize the network parameters, with a initial learning rate, $\eta[0] = 0.001$, and momentum, $m= 0.9$. In order to accelerate convergence, and maximize each networks' performance, an adaptive decaying learning rate, $\eta$, is used, as described in Equation (\ref{deyaing_learning_rate}).

\begin{equation}\label{deyaing_learning_rate}
\eta[\textnormal{epoch}] = \eta[\textnormal{epoch}-1] \cdot 0.01^{\frac{\textnormal{epoch}}{100}}
\end{equation}
\color{black}

%


\begin{figure}[!t]
	\centering
	\includegraphics[width=0.3\textwidth]{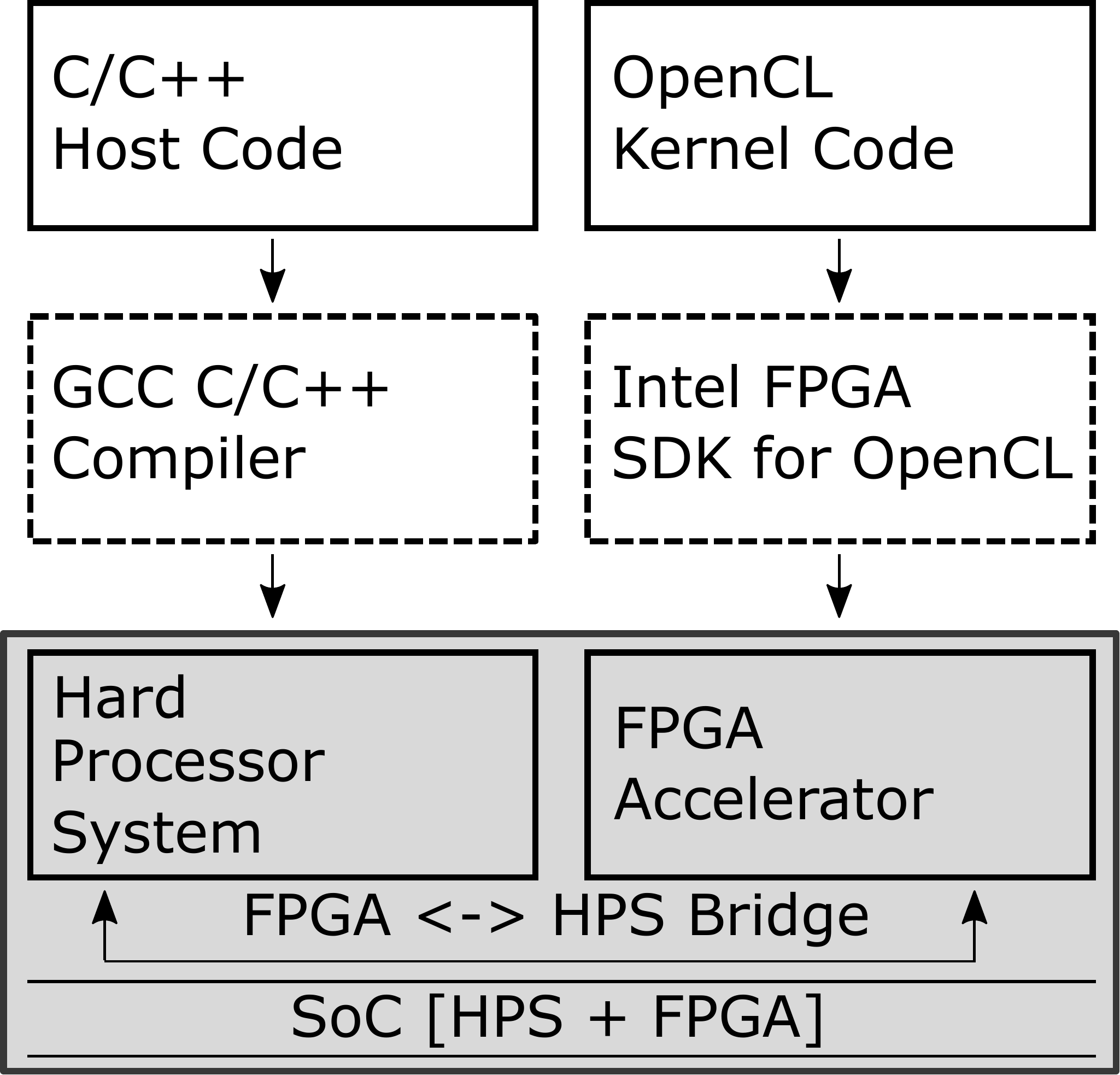}
	\caption{Top level flow diagram of the proposed network implemented on a SoC consisting of a processor describing the host controller and an FPGA to run OpenCL kernels.}
	\label{top_level_arch}
\end{figure}

\subsection{Hardware Architecture}
The developed hardware architectures consist of C++ host controllers and multiple OpenCL kernels, which are accelerated using either an FPGA or a GPU. For x86-based systems, OpenCL accelerated kernels using FPGAs typically reside on an FPGA development board, which is connected to a separate independent host system through the PCIe express interface~\cite{pipe_cnn}. For ARM-based systems, the FPGA is typically connected to a Hard Processor System (HPS) on a SoC through specialized bridges, as in the case of the Intel DE1-SoC development board that we used here. This allows the proposed networks to be run completely independently on the SoC without using a separate device for computation. 
The full top level flow diagram of our implemented FPGA-accelerated networks is presented in Figure (\ref{top_level_arch}). 

In addition to accelerating the targeted MNIST and CIFAR-10 networks on a FPGA development board, each network is also accelerated on a state-of-the-art \textit{Titan V GPU} to execute OpenCL kernels and an \textit{AMD Ryzen 2700X @ 4.10 GHz Overclocked (OC) CPU} to drive the operating system.

\color{black}
\begin{table*}[!t]\label{bigtable}
\centering
\caption{Implementation results obtained using the MNIST and CIFAR-10 datasets for GPU and FPGA accelerated networks. The Learning Time per Epoch and Inference Time per Image metrics are averaged over all recorded samples during 200 training epochs.}
\begin{adjustbox}{width=1\textwidth}
\begin{tabular}{rrrrrrrrr} 
\toprule
\multicolumn{1}{c}{\multirow{2}{*}{\textbf{Regularizer} }} & \multicolumn{2}{c}{\textbf{Total Kernel Power Usages (W)} } & \multicolumn{2}{c}{\textbf{Learning Time per Epoch (s)} } & \multicolumn{2}{c}{\textbf{Inference Time per Image (s)} } & \multicolumn{2}{c}{\textbf{Validation Accuracy (\%)} }  \\
\multicolumn{1}{c}{}                                       & \multicolumn{1}{l}{\textbf{FPGA}} & \textbf{GPU}            & \multicolumn{1}{l}{\textbf{FPGA}} & \textbf{GPU}          & \multicolumn{1}{l}{\textbf{FPGA}} & \textbf{GPU}           & \multicolumn{1}{l}{\textbf{FPGA}} & \textbf{GPU}        \\ 
\midrule
\multicolumn{9}{c}{MNIST}                                                                                                                                                                                                                                                                                   \\ 
\midrule
\multicolumn{1}{l}{No Regularizer} & 7.0 & 126.1 & 26.09 & 5.13 & 7.04E-05 &                        3.12E-05 & 98.70 & 98.54\\
Deterministic & 6.3 & 125.9 & 9.75 & 8.87 & 6.84E-06 & 9.71E-06 & 97.76 &                     97.94\\
Stochastic & 6.3 & 125.4 & 11.58 & 8.20 & 7.12E-06 & 9.92E-06 & 98.33 & 98.23                    \\ 
\midrule
\multicolumn{9}{c}{CIFAR-10}                                                                                                                                                                                                                                                                                \\ 
\midrule
No Regularizer  & 7.9 & 128.4 & 43.97 & 28.45 & 1.15E-02 & 5.09E-03 & 86.72 & 86.73\\
Deterministic   & 6.5 & 126.3 & 16.91 & 34.86 & 1.11E-03 & 1.63E-03 & 86.48 & 86.46\\
Stochastic      & 6.6 & 126.9 & 20.08 & 33.79 & 1.16E-03 & 1.66E-03 & 86.75 & 86.76\\
\bottomrule
\end{tabular}
\end{adjustbox}\label{tab1}
\end{table*}

\color{black}
\section{Implementation Results}

\begin{figure}[b!]
	\centering
	\includegraphics[width=0.5\textwidth]{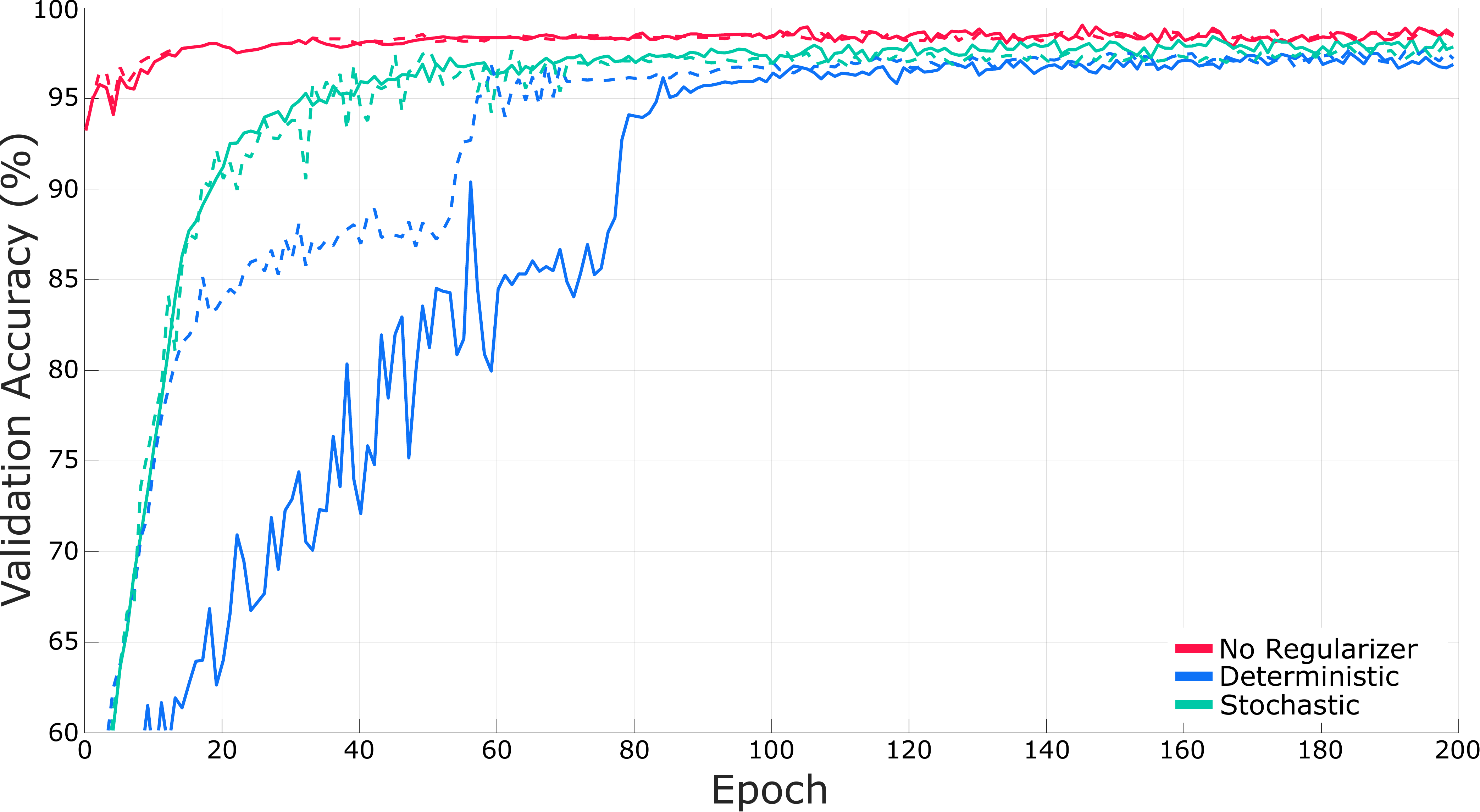}
	\caption{Validation Accuracy during training for FPGA- and GPU-accelerated permutation-invariant FCNN for the MNIST test set. Solid lines represent the validation error for networks accelerated using FPGA and dashed lines represent the validation error for networks accelerated using GPU.}
	\label{mnist_training_curve}
\end{figure}

\begin{figure}[b]
	\centering
	\includegraphics[width=0.5\textwidth]{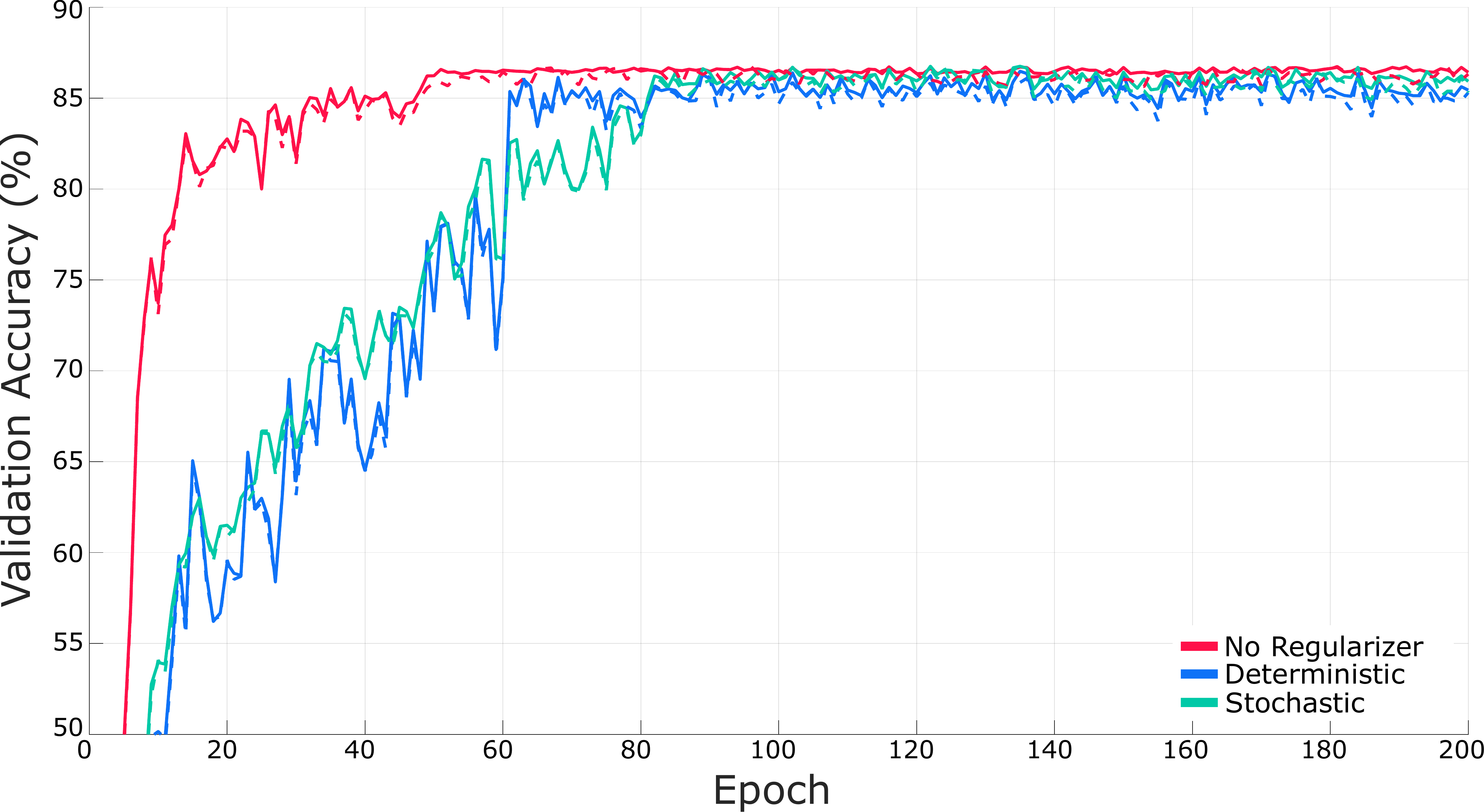}
	\caption{Validation Accuracy for each Epoch during training for FPGA- and GPU-accelerated VGG-16 CNN for the CIFAR-10 test set. Solid lines represent the validation error for networks accelerated using FPGA and dashed lines represent the validation error for networks accelerated using GPU.}
	\label{cifar_training_curve}
\end{figure}

In order to validate and investigate the performance of the proposed FPGA- and GPU-accelerated BNN architectures, the MNIST and CIFAR-10 datasets are used. 
To ensure a fair comparison, on account of the limited resource availability of the Intel DE1-SoC development board used, the batch size, $\Im$, was fixed to 4 for all networks. 
The validation accuracy for all developed networks over 200 training epochs is presented in Figures (\ref{mnist_training_curve}) and (\ref{cifar_training_curve}).

From Figures (\ref{mnist_training_curve}) and (\ref{cifar_training_curve}), it can be observed that both GPU- and FPGA-accelerated networks achieve very similar validation accuracy rates during learning. For all implementations, regularized networks require more training epochs to converge.

The variations in validation accuracy trends reported between platforms can be associated to the different initial weights generated using the He weight initialization technique.
Figures (\ref{mnist_training_curve}) and (\ref{cifar_training_curve}) also demonstrate that, networks employing stochastic and deterministic binarization techniques perform very similarly, compared to their base-line architectures employing no binary regularization techniques. For our FPGA-accelerated networks learning MNIST, the validation accuracy degrades by only 0.37\% (for stochastic) and 0.94\% (for deterministic), compared to no regularization. For our FPGA-accelerated networks learning CIFAR-10, a validation accuracy decrease of 0.24\% was observed for our network employing deterministic binarization, while our networks with stochastic binarization regularization showed a validation accuracy increase of 0.03\%. These findings are in good agreement with the software implementations of the binarized networks reported in~\cite{2015arXiv151100363C}.

To comprehensively compare the implemented FPGA- and GPU-accelerated networks, the total kernel power usages, learning time per epoch, inference time per image, and learning performances were determined and presented in Table~\ref{tab1}. The total kernel power usages were determined using the \textit{Post Place \& Route Estimator} for FPGA post-synthesis, and \textit{NVIDIA-SMI} for GPU. It was found that the power consumption of all FPGA-accelerated networks reduce by $>$16 times, compared to their GPU-accelerated counterparts. 

\color{black}
Despite this drastic reduction in power consumption, our deterministic and stochastic regularized FPGA-accelerated networks require similar training durations to their GPU-accelerated counterparts, which have much higher operation frequencies, compared to our utilized FPGA. As reported in Table~\ref{tab1}, our FPGA-accelerated permutation-invariant FC stochastic and deterministic BNNs for MNIST require 1.10$\times$ and 1.41$\times$ longer training intervals, respectively. Our FPGA-accelerated CNNs adopting the VGG-16 architecture accelerate learning by 2.06$\times$ and 1.68$\times$, respectively. These findings are in agreement with~\cite{PPM}, which investigates execution times for FC and CNN BNNs, and demonstrates that convolutional operations are accelerated to a greater extent than matrix multiplications, which are required for FC layers. 

When considering the inference time, all our FPGA-accelerated stochastic and deterministic regularized networks require shorter times to perform inference, compared to their GPU-accelerated counterparts. This is notable, considering our GPU-accelerated networks use the state-of-the-art \textit{Titan V GPU} to execute OpenCL kernels, while the limited resources available on the utilized FPGA creates a large bottleneck on the maximum synthesizable frequency, and thus limits the speed of our FPGA-accelerated networks. We believe the shorter inference times observed are mainly due to the binarized parameters during inference, which accelerate the required computations. This also explains why, when no regularizer is used, our GPU-accelerated implementations require shorter inference times than our FPGA-accelerated implementations. Modern FPGAs such as the Stratix® V GXA7 and Virtex-7 VX485T, used in other recent works~\cite{bcnn, pipe_cnn}, are expected to demonstrate even more significant improvements in speed during training and inference. This promises further inference acceleration for FPGA-based deterministically and stochastically binarized networks.

\section{Conclusion}
We designed and implemented various FC and convolutional BNN architectures using the high-level OpenCL framework. We then accelerated the developed networks on both GPUs and FPGAs. The performance, power, and learning/inference times of these network architectures were investigated. It was found that both FPGA-accelerated BNNs with deterministic and stochastic regularizers have reduced inference times on MNIST and CIFAR-10 by an order of magnitude, compared to the no-regularized FPGA case. They require $>$25\% shorter inference times than their GPU counterparts. Moreover, our FPGA-accelerated BNNs consumed less than 16$\times$ the power required by non-regularized GPU-accelerated networks. Finally, our BNNs achieved only slightly degraded validation errors on MNIST, and in some instances, outperformed our baseline non-regularized GPU-accelerated networks on CIFAR-10. In summary, our modular and scalable FC and CNN network architectures can be extrapolated to accelerate larger and more complex networks.
\color{black}
%

\bibliographystyle{IEEEtran}
\bibliography{DNN_BNN_BIB,SC}

\end{document}